\documentclass[conference]{IEEEtran}
\IEEEoverridecommandlockouts
\usepackage{float}
\usepackage{booktabs} 
\usepackage{subcaption}
\usepackage{cite}
\usepackage{amsmath,amssymb,amsfonts}
\usepackage{algorithmic}
\usepackage{graphicx}
\usepackage{textcomp}
\usepackage{xcolor}
\def\BibTeX{{\rm B\kern-.05em{\sc i\kern-.025em b}\kern-.08em
    T\kern-.1667em\lower.7ex\hbox{E}\kern-.125emX}}
\begin{document}

\title{Personalized Observation Normalization for Federated Reinforcement Learning in Simulation Environments with Heterogeneity  \\


\thanks{
\IEEEauthorrefmark{1}Corresponding author.

This paper has been accepted for publication at the International Joint Conference on Neural Networks (IJCNN) 2025.
}

}

\author{

\IEEEauthorblockN{Yiran Pang, Zhen Ni, and Xiangnan Zhong\IEEEauthorrefmark{1}}
\IEEEauthorblockA{
\textit{Department of Electrical Engineering \& Computer Science} \\
\textit{Florida Atlantic University} \\
Boca Raton, FL, USA \\
ypang2022@fau.edu, zhenni@fau.edu, xzhong@fau.edu
}

}

\maketitle

\begin{abstract}
Federated reinforcement learning (FedRL) enables multiple agents to collaboratively train a global policy without sharing raw data, making it ideal for privacy-sensitive applications. However, FedRL faces challenges in heterogeneous environments where differing state-transition dynamics lead to non-identical input distributions and imbalanced parameter updates during aggregation. Therefore, this paper develops a personalized observation normalization (PON) method, allowing each agent to locally normalize raw state inputs using a continuously updated running mean and variance. This design ensures consistent scaling of local feature without overshadowing across agents during aggregation. Furthermore, we demonstrate that sharing normalization parameters across agents is ineffective due to the diverse local input distributions, which highlights the necessity of personalized statistics. Experiments on heterogeneous MuJoCo tasks show that our developed PON accelerates training and achieves superior performance compared to baseline methods. 
\end{abstract}


\section{Introduction}

\begin{figure*}[htbp]
    \centering
    \subfloat[Each agent interacts with a locally distinct environment in FedRL]{
        \includegraphics[width=0.45\textwidth]{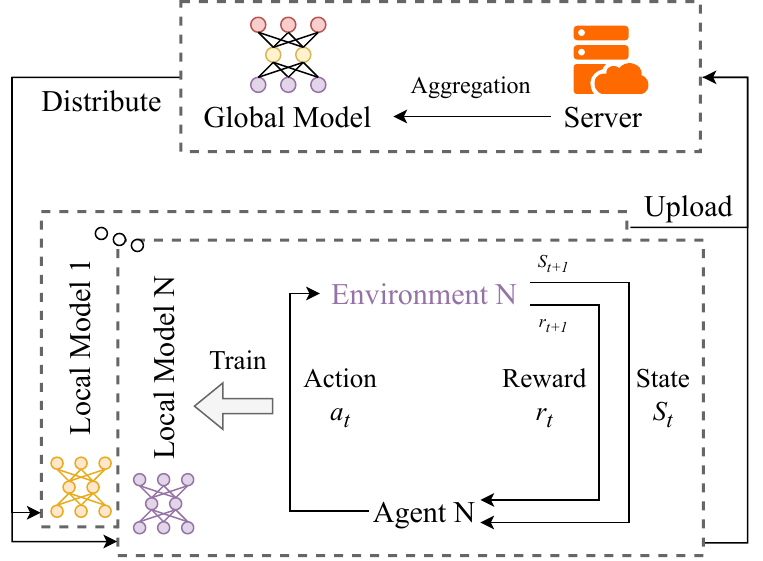}
        \label{fig:framework}
    }
    \hspace{0.02\textwidth} 
    \subfloat[Heterogeneous environment statistics]{
        \includegraphics[width=0.42\textwidth]{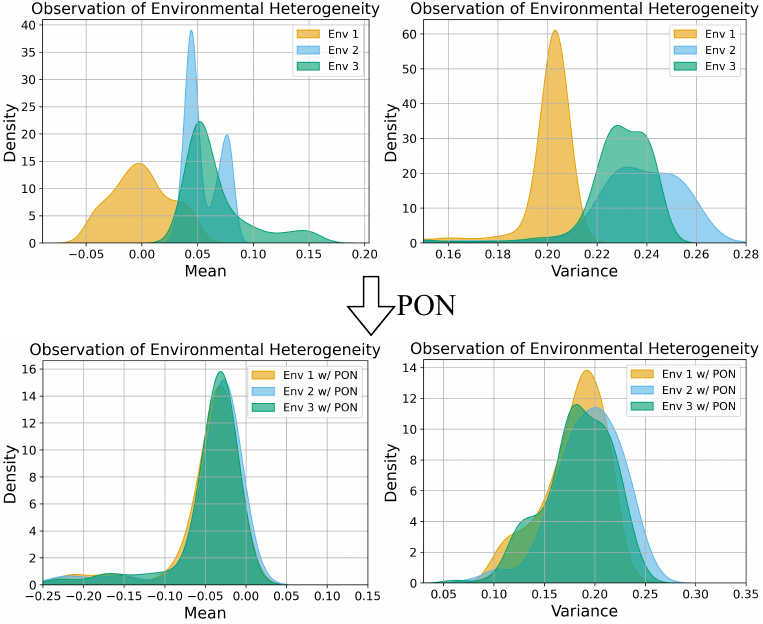}
        \label{fig:distribution}
    }
    \caption{
        (a) Illustration of the FedRL system. The global model is distributed to and aggregated from local models, with each local model interacting with its distinct environment. (b) Observation distributions from three agents in heterogeneous HalfCheetah environments. The upper plot demonstrates the density of observation values across agents, highlighting the mean and variation in input distributions caused by differing state-transition dynamics. The lower plot shows the effect of applying personalized observation normalization (PON), which aligns the distributions across agents.
    }
    \label{fig:combined}
\end{figure*}

Reinforcement learning (RL) is centered on enabling agents to interact with their environment to maximize cumulative rewards. While RL has achieved remarkable success \cite{10651449, 10847305}, it faces critical challenges in practical applications, particularly in efficiently exploring large state-action spaces. To address this, distributed RL \cite{10651403, fernandez2024simion} and parallel RL methods \cite{das2023microgrid, 10689197} have been introduced, enabling agents to share information and collaborate. However, these approaches often rely on centralized data collection, leading to significant communication overhead. Recently, federated learning (FL) has emerged as a promising distributed approach that enables model training directly on decentralized edge devices, providing enhanced privacy protection and improved communication efficiency \cite{10650184, FLCC}. 

In addition, federated reinforcement learning seamlessly combines FL and RL to address the challenges of communication overhead and data privacy in distributed decision-making frameworks. By enabling geographically dispersed agents to perform multiple local learning iterations before periodically transmitting intermediate results (such as neural network gradients or policy parameters) to a central server, FedRL significantly reduces communication frequency and enhances overall efficiency. This approach not only minimizes overhead but also ensures data privacy by eliminating the need for raw data transmission, making it scalable and adaptable to diverse applications. Therefore, FedRL has been widely used in various domains, including vehicular operations, energy management, and IoT systems \cite{qiu2023federated, pinto2023federated, PANG2025129550}, where privacy and efficient coordination are critical. 

Despite the success, FedRL still faces significant challenges arising from heterogeneous data distributions. These challenges stem from agents operating in diverse environments, which result in variations in the data they collect. For example, in vehicular networks \cite{Dosovitskiy17}, differences in road infrastructure, sensor hardware, and operational contexts generate statistically distinct data streams, resulting in the non-independent and identically distributed (non-i.i.d.) data problem \cite{li2020federated, hsieh2020non, chen2022calfat}. Such distribution drifts destabilize the FedRL training process, causing conflicting gradient updates during model aggregation and ultimately degrading the performance of the global policy.

To solve this issue, personalized FedRL \cite{chen2024review} methods have emerged as a promising solution. The personalization, introduced by Fallah et al. \cite{fallah2020personalized}, aims to tailor models to the heterogeneous data distributions of different users. It achieves this by learning a shared initial model while allowing each user to retain specific parameters, enabling personalized local models within a collaborative framework. Nadiger et al. \cite{nadiger2019federated} introduced a FedRL method combining Deep Q-Network to accelerate personalized model training, addressing heterogeneity through data-type-based aggregation to improve model accuracy. Qi et al. \cite{qi2021federated} proposed to address non-i.i.d issues by enhancing policy generalization in FedRL. Wu et al. \cite{wu2022dpfed} proposed DPFed, a personalized federated learning framework using deep RL to identify client relationships and address heterogeneity under non-i.i.d data, enabling collaborative learning among similar clients and improving model convergence. Jin et al. \cite{jin2022federated} employ a personalized FedRL method to add an embedding layer in the neural network for each agent in a heterogeneous environment, enabling local policies to be personalized during the training process. Xiong et al. \cite{xiong2024personalized} proposed perFedDC for addressing environmental heterogeneity by balancing personalization and shared experience. However, these methods primarily focused on adapting to heterogeneous data distributions at the model level, without considering the input distribution shifts caused by environmental heterogeneity.

Normalization methods are commonly used to mitigate distribution shifts in machine learning. Among them, FedBN~\cite{li2021fedbn} addresses feature heterogeneity in supervised federated learning by preserving local batch normalization (BN) statistics across clients. However, BN relies on stable batch-wise statistics and is often unsuitable for RL settings, where observation distributions are inherently non-stationary due to ongoing policy updates~\cite{bhatt2024crossq}. Observation normalization (ON) provides a direct approach to stabilizing training by standardizing input features, addressing input distribution changes \cite{engstrom2020implementation, stable-baselines3}. Conventional ON implementations, which rely on pre-computed statistics or manual specifications when observation boundaries are known \cite{maze_rl_observation_normalization}, have proven effective in centralized RL. However, these static strategies are insufficient for heterogeneous FedRL, where agents operate in distinct environments and policies evolve over time.  

To tackle these challenges, we propose a personalized observation normalization (PON) approach tailored for FedRL systems. As illustrated in Fig. 1(a), the global model of FedRL is distributed to multiple local models, each interacting with its own unique environment. The results from these local models are then aggregated to update the global model. Heterogeneous input distributions across agents can cause inconsistencies during federated aggregation, resulting in biased global models and slower convergence. Our designed PON method addresses these issues by integrating observation normalization into the training process of each agent. Each agent retains its own mean and variance statistics locally, capturing the dynamics of its specific environment without sharing these statistics globally. This method ensures that the interference from other heterogeneous environments is prevented, allowing the system to effectively adapt to unique state distributions. As depicted in Fig. 1(b), PON scales and aligns input distributions from diverse environments, thereby facilitating better knowledge sharing and improving training efficiency in each agent.

In summary, our contributions are as follows:

\begin{itemize}
    \item We propose the PON method to address the challenge of input distribution heterogeneity in FedRL. 
    Specifically, this method enables dynamical updates on observation statistics for each agent during local training by maintaining time-step-wise running mean and variance. This adaptive normalization aligns with the evolving state distributions of individual agents. Meanwhile, this design retains local statistical parameters and excludes them from global aggregation, ensuring that each agent's observations are normalized specifically for its unique environment without interference from other heterogeneous environments. Additionally, we investigate the limitations of sharing normalization parameters in heterogeneous environments and highlight the critical need for personalized normalization.
    
    \item We develop the FedRL algorithm based on the proximal policy optimization (PPO) framework, which integrates federated averaging (FedAvg) with PPO to facilitate privacy-preserving collaborative training among distributed agents. By periodically aggregating the parameters of both the policy network and the value function network, FedRL-PPO preserves the stability and robustness of PPO while leveraging the collective experience from the federated environment to enhance learning efficiency. Experimental results in heterogeneous environments demonstrate that FedRL-PPO surpasses the standalone PPO baseline, achieving superior convergence speed and final performance across all participating agents. 
    
    \item We construct multiple MuJoCo environments with distinct morphological variations to simulate realistic heterogeneity. These environments are designed to reflect the diverse conditions under which FedRL agents may operate in realistic scenarios. Experimental results demonstrate that PON enhances the training speed and final performance of FedRL in heterogeneous environments, validating its effectiveness in addressing environmental heterogeneity.
\end{itemize}

The remaining paper is organized as follows: Section II provides an background of RL, PPO, and the FedRL process. Section III outlines the heterogeneous input distribution problem and propose the PON and FedRL-PPO algorithms. Section IV presents the experimental setup, the results across three environments, and the failure of sharing the normalization parameters. Finally, Section V concludes the paper and discusses potential directions for future research.

\section{Background}
\subsection{Reinforcement Learning and Proximal Policy Optimization}
RL problems are typically formulated as markov decision processes (MDPs), defined by the tuple $(\mathcal{S}, \mathcal{A}, P, R, \gamma)$, where $\mathcal{S}$ denotes the state space, $\mathcal{A}$ is the action space, $P(s'|s,a)$ represents the state-transition probability distribution, $R(s,a)$ is the reward function, and $\gamma \in (0,1)$ is the discount factor. At each timestep $t$, an agent observes a state $s_t \in \mathcal{S}$, selects an action $a_t \in \mathcal{A}$ based on its policy $\pi_{\theta}(a_t|s_t)$ parameterized by $\theta$, and receives a scalar reward $R(s_t,a_t)$ while transitioning to a new state $s_{t+1}$. The objective is to maximize the expected cumulative discounted reward, $J(\theta) = \mathbb{E}_{\tau \sim \pi_\theta}\big[\sum_{t=0}^\infty \gamma^t R(s_t,a_t)\big]$, where $\tau$ represents a trajectory sampled under $\pi_\theta$.

Among RL methods, proximal policy optimization (PPO) \cite{schulman2017proximal} has gained widespread recognition due to its empirical stability and ease of implementation. PPO improves upon standard policy gradient methods by introducing a clipped surrogate objective that prevents large policy updates, thereby enhancing training stability. PPO achieves superior sample efficiency and robustness by encouraging policy improvement and constraining deviation from the old policy, making it a popular choice for various applications.

\begin{figure*}[!ht]
    \centering
    \includegraphics[width=0.75\textwidth]{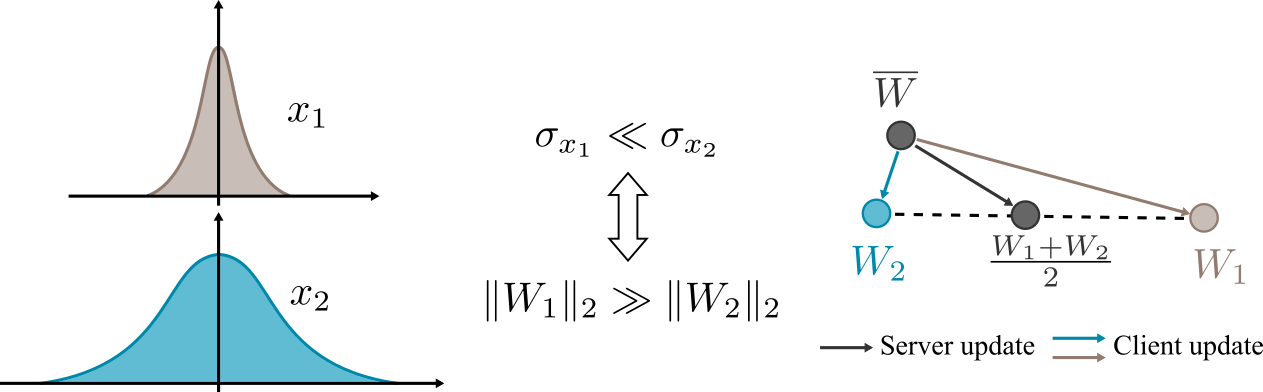} 
    \caption{Illustration of the issue where a model with a smaller norm is overshadowed by one with a larger norm during aggregation, as discussed in FL scenario \cite{du2022rethinking}.}
    \label{fig:external_shift}
\end{figure*}

\subsection{Federated Reinforcement Learning}  
FedRL aims to collaboratively train a global model from multiple decentralized agents without explicitly sharing their raw data. Consider $n$ agents, each interacting with an independent environment. Suppose we have $n$ MDPs $\{ \mathcal{M}_i \}_{i=1}^{n}$, each defined as $\mathcal{M}_i = \langle \mathcal{S}, \mathcal{A}, P_i, R, \gamma \rangle$, sharing a common state space $\mathcal{S}$, action space $\mathcal{A}$, and reward function $R$, but differing in state-transition dynamics $P_i$. At each agent $i$, the agent locally aims to improve its policy $\pi_{\theta}$ based on trajectories generated from $P_i$. The overarching goal of FedRL is to find a policy $\pi_{\theta}$ that performs well across all heterogeneous environments, achieving a form of cross-environment generality.

Under privacy constraints, agents cannot exchange raw trajectories directly. Instead, each agent locally updates its policy function for multiple times, and then the server average these $n$ functions of all agents. To improve the communication efficiency, the local updates are performed multiple times between two communications.

A policy-based FedRL formulation can be expressed as:
\begin{equation}\label{eq:fedrl_objective}
\max_{\theta} \frac{1}{n} \sum_{i=1}^{n} J_i(\theta) = \frac{1}{n}\sum_{i=1}^{n} \mathbb{E}_{\tau_i \sim \pi_{\theta}}\left[\sum_{t=0}^{\infty}\gamma^{t} R(s_{t}^{i}, a_{t}^{i})\right],
\end{equation}
where $\tau_i$ denotes trajectories sampled under policy $\pi_{\theta}$ in environment $i$.

In this work, we parameterize the policy network $\theta$ on agent $i$  by $(W, b)$, the forward operation can be represented as:
\begin{equation}
\label{eq:linear_layer}
y_i = W_i x_i + b.
\end{equation}
where $W$ and $b$ denote the weight matrix and bias vector, respectively. Here, the input $x_i$ is sampled from the local state distribution:
\begin{equation}
x_i \sim d_{\pi_\theta}^i(s).
\end{equation}
where $d_{\pi_\theta}^i(s)$ reflects the state distribution induced by the local policy $\pi_\theta$ at agent $i$. Over multiple rounds of training, the $i$-th agent performs the following local update:
\begin{equation}\label{eq:local_update}
W_i \leftarrow W_i - \alpha \nabla_{W_i} L_i(W_i,b_i),
\end{equation}
where $L_i(\cdot)$ is the local objective related to the policy gradient loss and $\alpha$ is a learning rate. Each agent thus obtains locally adapted parameters $W_i$ and $b_i$ that reflect the statistics and data distribution in its own environment. 

After performing multiple local updates, each agent uploads its locally adapted parameters $\overline{W}$ and $\bar b$ to the server. The server then aggregates the parameters across all agents using FedAvg  \cite{zhou2021communication}:
\begin{equation}
\label{eq:federated_avg_agg}
\overline{W} \leftarrow \frac{1}{n} \sum_{i=1}^{n} W_i, \quad \bar b \leftarrow \frac{1}{n} \sum_{i=1}^{n} b_i
\end{equation}
This aggregated global model is subsequently redistributed to all agents, enabling their local models to benefit from the collective learning across all agents:
\begin{equation}
\label{eq:federated_avg_afteragg}
W_i \leftarrow \overline{W}, \quad b_i \leftarrow \overline{b}, \quad \; \forall \, i\in n.
\end{equation}
This iterative process of local updates and global aggregation continues until convergence is achieved or a predefined number of training episodes is reached.

\section{Personalized Observation Normalization\\ for FedRL}
\subsection{Problem Formulation}\label{sec:AON}

In FedRL, heterogeneous input distributions arise due to the distinct state-transition dynamics $P_i$ in each agent’s local environment. Such heterogeneity cause each agent to learn distinct scaling factors in their parameters. The phenomenon slows down the convergence of the global model in FedRL, since the feature distribution varies after aggregation and neurons have to keep adapting the updated feature distribution.

To describe why the aggregation step is harmed by heterogeneous input distributions, we consider a toy example. Let $x_i$ be the input features at agent $i$, with mean $\mu_{x_i}$ and variance $\sigma_{x_i}^2$:
\begin{equation}\label{eq:mean_var_x}
\mathbb{E}[x_i] = \mu_{x_i}, \quad \mathrm{Var}(x_i) = \sigma_{x_i}^2.
\end{equation}
Consider a fully connected layer as Eqn. (\ref{eq:linear_layer}). If $W_i$ acts as a linear transform that primarily scales the input features, then the output variance can be approximated by
\begin{equation}\label{eq:var_y_simplified}
\mathrm{Var}(y_i) \approx \|W_i\|_2^2 \sigma_{x_i}^2,
\end{equation}
where $\|\cdot\|_2$ denotes the Euclidean norm. 

Over multiple communication rounds, federated aggregation drives the global model toward a stable equilibrium, typically a suboptimal solution due to the heterogeneity of agent environments. This behavior is theoretically supported by \cite{jin2022federated}, which formalizes the convergence of the global model under an averaged MDP constructed from the collective dynamics of all agents. At this equilibrium, we posit that the global model's output feature variance stabilizes to a specific value $\Delta y$. Since the global model operates on the averaged MDP, to ensure statistical consistency between the agents and the global model, the output variance of the agent $i$ should align with the global variance $\Delta y$, expressed as:
\begin{equation}\label{eq:delta_y_equilibirum}
\mathrm{Var}(y_i) \approx \Delta y, \quad \forall i.
\end{equation}

Combining \eqref{eq:var_y_simplified} and \eqref{eq:delta_y_equilibirum}, we derive:
\begin{equation}\label{eq:w_norm_relation_no_zero_mean}
\|W_i\|_2 \approx \frac{\sqrt{\Delta y}}{\sigma_{x_i}}.
\end{equation}

For two sets of inputs $x_1$ and $x_2$, if their feature deviations have significant discrepancies, i.e., $\sigma_{x_1} \ll \sigma_{x_2}$, then we can derive that $\|W_1\|_2 \gg \|W_2\|_2$ by Eqn. (\ref{eq:w_norm_relation_no_zero_mean}). Thus, the
essence of heterogeneous input distributions describes the shift of model parameter’s norm. Considering these two sets of weights belonging to two local models involved in FedRL, then the contribution of $W_1$ will be obliterated by $W_2$ as shown in Fig. \ref{fig:external_shift}, and the same effect applies to the bias, which is more related to $\mu_{x_i}$.

Hence, the heterogeneous input distributions affects the balance of contributions in the federated aggregation and can potentially bias the global model towards certain agents’ distributions.

\subsection{FedRL with Personalized Observation Normalization}

This section proposes FedRL with an PON approach. Initially, all agents start with identical policy parameters. Normalizing the observed data helps constrain the weight norms, thereby mitigating the detrimental effects caused by heterogeneous input distributions. 

\begin{figure}[!bhtp]
    \centering
    \includegraphics[width=0.48\textwidth]{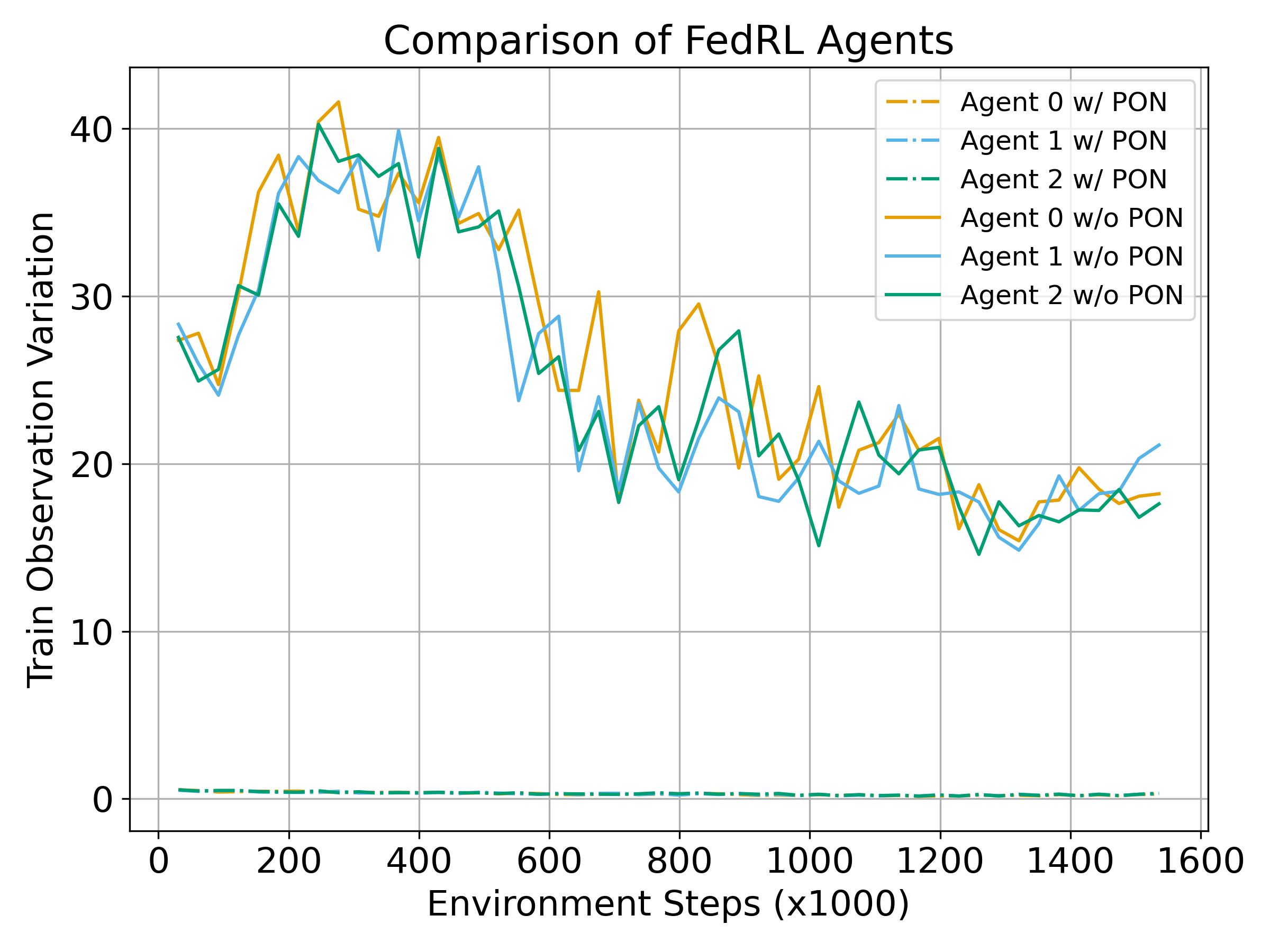} 
    \caption{The train observation variation over environment steps for agents with and without PON in heterogeneous HalfCheetah. The figure illustrates that the input variation fluctuates due to evolving policies and environment heterogeneity, while PON effectively reduces variation.}
    \label{fig:vs}
\end{figure}

As shown in Fig. \ref{fig:vs}, the input distributions of agents evolve dynamically along with their policies as training progresses. This evolution leads to notable changes in the input distribution variance over time. Furthermore, during each federated aggregation step, the heterogeneity of agent environments causes sharp shifts in input variation, further amplifying inconsistencies across agents. These fluctuations can undermine stable training and consistent aggregation in the FedRL process.

To address this critical challenge, we modify the variance-calculating algorithm proposed by Chan et al.\cite{chan1982updating} to fit the FedRL environment. We adjust the original accumulated mean and variance calculation to record the observed distribution continuously, incorporate newly arrived samples, and deploy to every agent in the FedRL framework. Our reparameterized update procedure is detailed below.

Let \( C_{t} \), \( \mu_{t} \), and \( \sigma^2_{t} \) represent the cumulative number of observations, the running mean, and the running variance of all observations, respectively. At time \( t \), a new batch of observations with \( C_b \) data is obtained, with batch mean \( \mu_b \) and batch variance \( \sigma_b^2 \). We first update the total count:

\begin{equation}
C_t \leftarrow C_{t-1} + C_b.
\end{equation}

The running mean is updated using the difference between the batch mean and the previous running mean:
\begin{equation}
\Delta = \mu_b - \mu_{t-1}, \quad \mu_t \leftarrow \mu_{t-1} + \Delta \cdot \frac{C_b}{C_t}.
\end{equation}

To update the running variance, define:
\begin{equation}
m_{t-1} = \sigma^2_{t-1} C_{t-1}, \quad m_b = \sigma_b^2 C_b.
\end{equation}
The updated running variance is computed as:
\begin{equation}
\sigma_t^2 \leftarrow \frac{m_{t-1} + m_b + \frac{\Delta^2 C_{t-1} C_b}{C_t}}{C_t}.
\end{equation}

After obtaining \( C_t \), \( \mu_t \), and \( \sigma_t^2 \), each new observation \( x \) in the batch at time \( t \) is normalized as:
\begin{equation}
\tilde{x} = \frac{x - \mu_t}{\sqrt{\sigma_t^2 + \epsilon}},
\end{equation}
where \( \epsilon \) is a small positive constant introduced to prevent division by zero.

It is noteworthy that in the PON approach, statistical parameters ($\mu$, $\sigma$) are not shared among agents. Instead, each agent maintains and utilizes its local statistical parameters for normalizing local inputs. This personalized scheme ensures that agents' normalization remains unaffected by other heterogeneous environments' statistical variations. Since the statistics are updated via incremental formulas and stored as low-dimensional vectors, the computational and memory overhead of PON is negligible.

\subsection{Failure of Share Normalization Parameters}
\begin{figure}[!bhtp]
    \centering
    \includegraphics[width=0.48\textwidth]{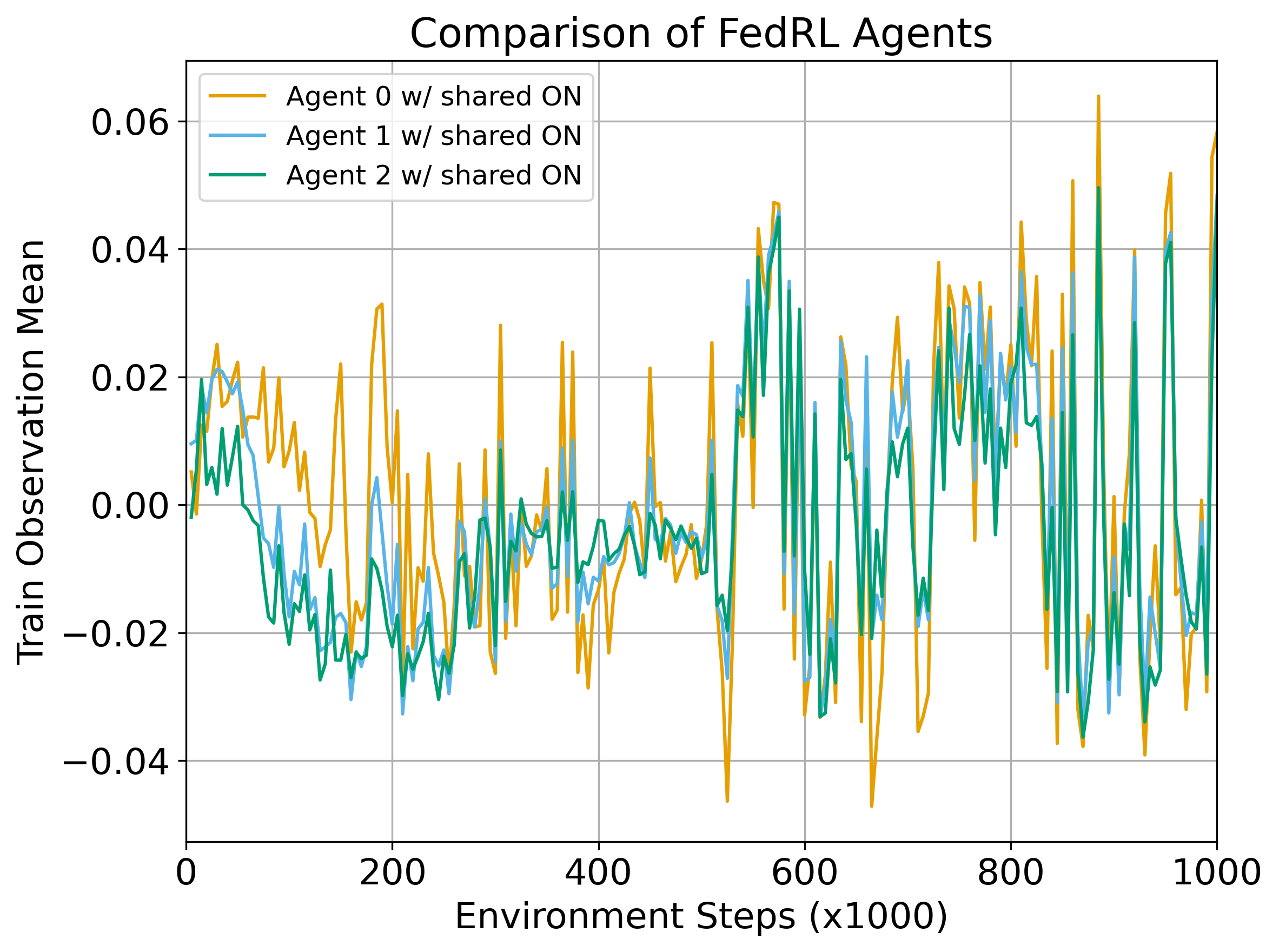} 
    \caption{The mean observation trajectories over training steps for three agents using ON with normalization parameters shearing (shared ON). The shared parameters lead to inconsistent observation scaling across heterogeneous environments, causing significant fluctuations in the observation mean.}
    \label{fig:FedAON}
\end{figure}

A straightforward idea for applying PON to a federated variance is to aggregate the local running statistics (i.e., mean and variance) across all agents. In this approach, shared ON, each agent $i$ computes its local mean, $\mu_t^{(i)}$, and variance, $\sigma_t^{2(i)}$, from its own observation batch at each training round $t$. After a few local updates, each agent sends $\mu_t^{(i)}$ and $\sigma_t^{2(i)}$ to the server for aggregation. The server then applies a averaging scheme (analogous to Eqn. \eqref{eq:federated_avg_agg}) to obtain the global statistics:
\begin{equation}\label{eq:FedAON_agg_stats}
    \bar{\mu}_t \leftarrow \frac{1}{n} \sum_{i=1}^{n} \mu_t^{(i)}, 
    \quad
    \bar{\sigma}_t^2 \leftarrow \frac{1}{n} \sum_{i=1}^{n} \sigma_t^{2(i)},
\end{equation}
and broadcasts $\bar{\mu}_t$ and $\bar{\sigma}_t^2$ back to all agents.

Once these aggregated global statistics are received, each agent locally normalizes its observations $x_i$ sampled from its environment distribution $x_i \sim d_{\pi_\theta}^i(s)$ as:
\begin{equation}
\label{eq:fedaon_normalize}
    \tilde{x}_i = \frac{x_i - \bar{\mu}_t}{\sqrt{\bar{\sigma}_t^2 + \epsilon}}.
\end{equation}

While the shared ON allows all agents to share and leverage global statistics, it fails in heterogeneous environments. Specifically, each agent’s local environment induces distinct observation distributions $d_{\pi_\theta}^i(s)$ due to environmental heterogeneity. Thus, forcing a single pair of mean and variance parameters across all agents can lead to substantial deviations from standard-normalized inputs. As illustrated in Fig.~\ref{fig:FedAON}, this mismatch amplifies the variability of the normalized observations.

\begin{figure*}[htbp]
    \centering
    \includegraphics[width=\textwidth]{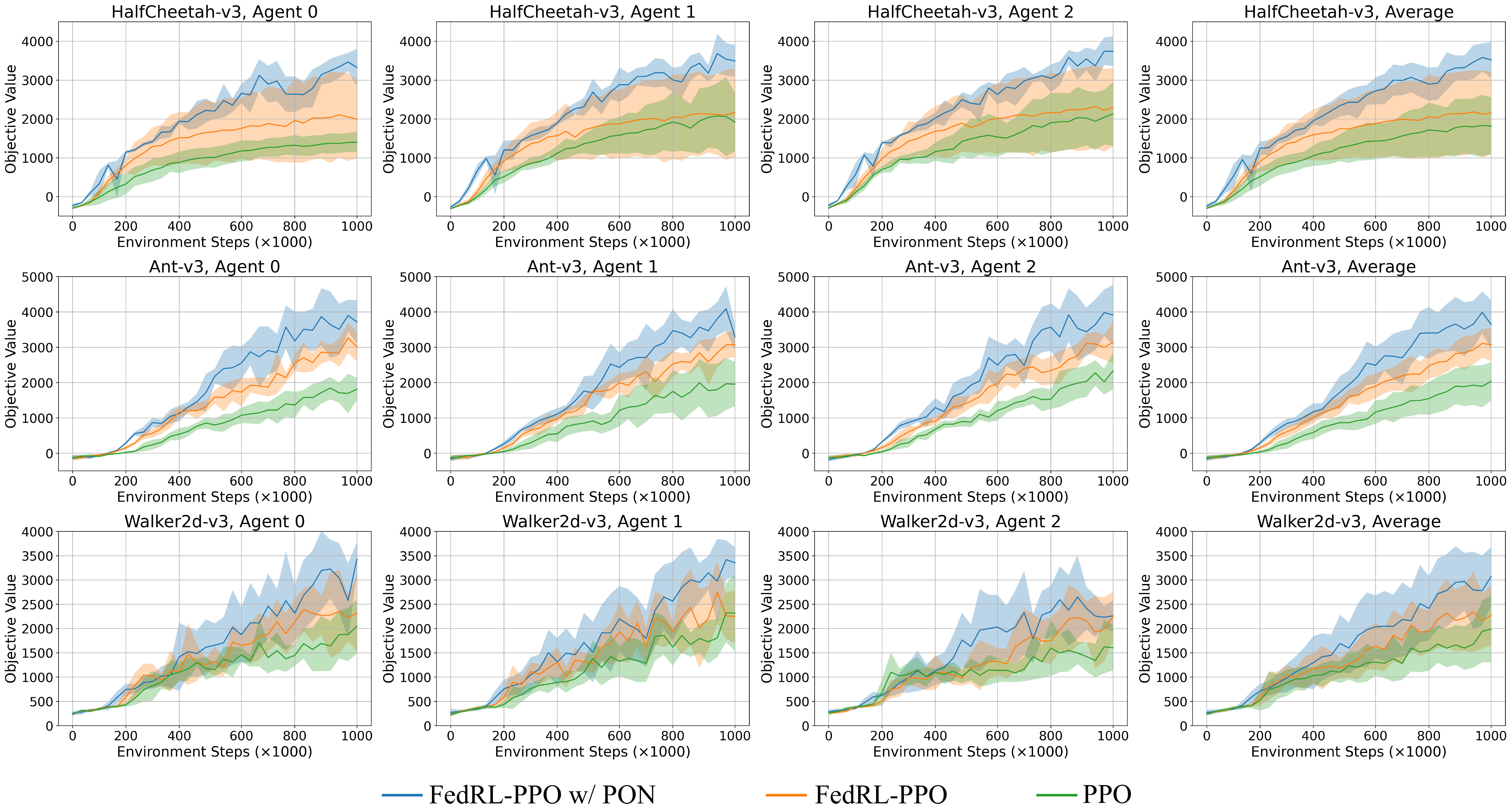} 
    \caption{Performance comparison of independent learning PPO, FedRL-PPO (PPO with FedAvg for Actor and Critic), and FedRL-PPO with PON in three Mujoco environments. Each environment includes three heterogeneous settings and the average performance of agents. The shaded area represents the standard deviation. Objective value is plotted against the number of environment steps. All experiments were repeated with five different random seeds. FedRL with PON achieves faster convergence and higher final performance across all settings.}
    \label{fig:result}
\end{figure*}

\begin{table*}[htbp]
\centering
\caption{Performance comparison of different methods across several heterogeneous environments. The value represents the average and standard deviation of the best performance of three agents in five repeated experiments.}
\label{tab:performance_comparison}
\begin{tabular}{lccc}
\toprule
Method & HalfCheetah & Ant & Walker2d \\
\midrule
PPO & $1923.08 \pm 828.42$ & $2168.26 \pm 529.07$ & $2304.43 \pm 571.84$ \\
FedRL-PPO & $2239.04 \pm 1098.35$ & $3370.11 \pm 386.52$ & $2931.00 \pm 646.13$ \\
\textbf{FedRL-PPO w/ PON} & $\mathbf{3938.03 \pm 288.82}$ & $\mathbf{4354.00 \pm 517.79}$ & $\mathbf{3532.11 \pm 403.20}$ \\
\bottomrule
\end{tabular}
\end{table*}

\subsection{FedRL-PPO Algorithm}
We develop FedRL-PPO algorithm building upon the PPO framework. This algorithm incorporates federated averaging to enable privacy-preserving collaborative training across distributed agents. The training process follows a cyclical pattern: First, each agent interacts with its local environment to collect trajectory data and computes policy gradients independently. These agents then perform local updates on their policy and value function networks using the collected data. After a predetermined number of local training epochs, agents upload their network parameters to a central server while keeping their raw environment interaction data private. The server aggregates these parameters by averaging, creating updated global models that are subsequently distributed back to all participating agents. This cycle of local training, parameter sharing, and global model synchronization repeats throughout the training process, allowing agents to benefit from collective learning while maintaining data privacy. Unlike existing policy averaging methods \cite{jin2022federated} that only aggregate parameters of the policy network, our approach simultaneously uploads and aggregates both the policy network and value function network. This synchronization ensures consistent policy-value alignment during distributed training.

\section{Experiments}
\subsection{Experimental Setup}

\textbf{Heterogeneous environment settings.} 
We conduct our experiments in a FedRL framework with three MuJoCo \cite{towers2024gymnasium} environments: HalfCheetah-v3, Ant-v3, and Walker2d-v3. Under the federated setting, each agent spawns an agent with a unique combination of environment’s morphology, creating heterogeneous training conditions.

\subsubsection{HalfCheetah}
We alter the capsule lengths of the front and back thighs. We uniformly sample ``bthigh'' length from $[0.10,\,0.19]$ and ``fthigh'' length from $[0.10,\,0.17]$. Each local agent employs a different combination of thigh lengths.

\subsubsection{Ant}
We group the legs into front and back pairs, applying a scale factor from $[0.8,\,1.2]$ to each pair. This preserves partial symmetry, both front legs have identical scaling and both back legs share another, but creates morphological diversity across agents.

\subsubsection{Walker2d}
We synchronize left and right legs by sampling a single length for both thighs and another for both shins. The thigh length may lie in $[0.35,\,0.55]$ and the shin in $[0.4,\,0.6]$. This ensures left-right balance while still allowing inter-agent variation.

Across all three environments, these morphological randomizations serve to replicate realistic discrepancies (e.g., manufacturing tolerances or assembly differences). 

\textbf{Baseline.} 
The point of PON is to enhance the performance of FedRL by aggregating and sharing knowledge across heterogeneous agents while maintaining data privacy. To verify the efficacy of PON, we establish two configurations for experimental comparison: one is vanilla PPO, where each agent trains a standalone PPO policy without any parameter sharing; the second is FedRL-PPO implementing parameter averaging separately on the policy network and value function network without normalization. This baseline demonstrates the benefits of parameter sharing while highlighting the potential shortcomings in handling heterogeneous environments. By comparing FedRL-PPO with PON against these baselines, we aim to showcase its ability to achieve superior performance through more sophisticated knowledge integration and policy optimization.

\textbf{Impliment details.} 
We implement the FedRL framework and PON using PyTorch \cite{imambi2021pytorch} and Tianshou \cite{tianshou}. We include the use of three heterogeneous agents in the FedRL framework. Each agent employs a fully connected MLP with two hidden layers (64 units each) and Tanh activations, outputting the mean of a Gaussian distribution with variable standard deviations, as described in \cite{schulman2017proximal}. The batch size is set to 64, the learning rate to $3 \times 10^{-4}$, and the discount factor $(\gamma)$ to 0.99.

\subsection{Performance Evaluation}
This section analyzes the performance of the PON with the FedRL learning method. We first justify that the PON helps accelerate training in individual environments in the heterogeneous setting and then compare our methods with Baselines regarding final performance.

To figure out the impact of heterogeneity on aggregation in FedRL, we compare the average performance in three heterogeneous Mujoco environments, as shown in Fig. \ref{fig:result}. The results indicate that compared to INDL and FedRL-PPO, the proposed FedRL-PPO with PON demonstrates faster convergence speed and higher final performance across all environments. Taking HalfCheetah-v3 as an example, the PON method accelerates each agent's learning process during the initial training phase. As training steps increase, the PON curve remains above other baselines for most periods and shows higher objective values in later stages of training. In contrast, FedRL-PPO, while outperforming INDL, which does not exchange information with other agents, struggles to maintain fast improvements in the training process due to heterogeneous input distribution. 

Further observation of numerical comparisons in Table \ref{tab:performance_comparison} reveals that FedRL-PPO with PON exhibits the most average performance improvement across three heterogeneous environments. Specifically, in HalfCheetah-v3 and Ant-v3 environments, average target values improved by several hundred to over a thousand points compared to baseline methods. The Walker2d-v3 environment also saw considerable improvements in convergence speed and final scores. This validates that PON can effectively coordinate FedRL learning processes under heterogeneous scenarios by considering individual environment differences while sharing information. This leads to superior performance in convergence metrics. In summary, experiments demonstrate that PON has the advantages of rapid convergence and high final results within heterogeneous FedRL settings.

\subsection{Effect of Share Normalization Parameters}
\begin{figure}[htbp]
    \centering
    \includegraphics[width=0.4\textwidth]{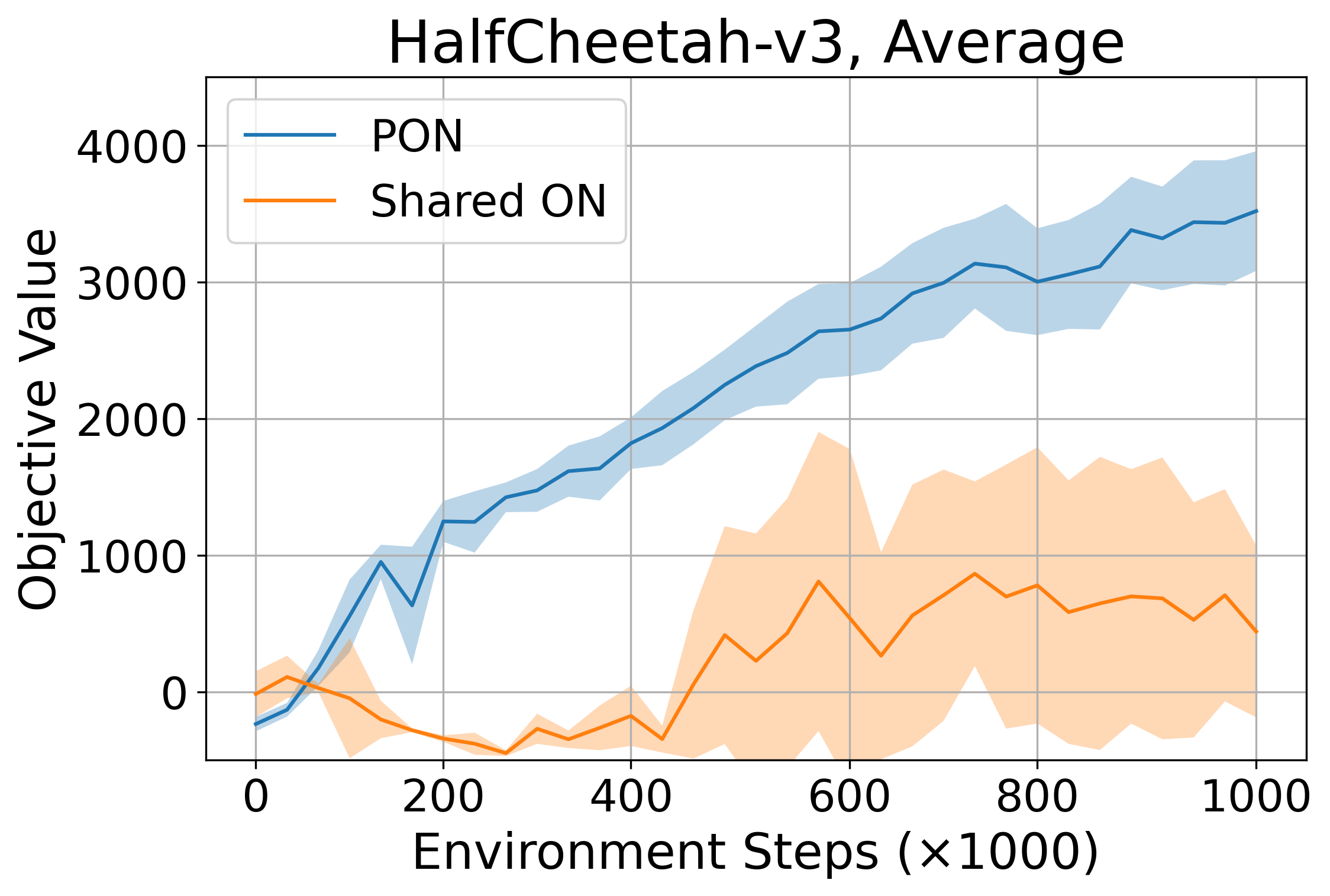} 
    \caption{Illustrates the FedRL training curves of PON and its shared-parameter variant, shared ON, on the heterogeneous HalfCheetah-v3 task. The solid line and shaded area represent the average and standard deviation of the objective value of three agents.}
    \label{fig:FedAONResult}
\end{figure}

Fig.~\ref{fig:FedAONResult} compares the average target value changes when training with PON and shared ON in the HalfCheetah-v3 environment. Due to sharing global normalization parameters in a heterogeneous environment, shared ON exhibits notable fluctuations during training. These performance discrepancies highlight the limitations of sharing global normalization parameters across heterogeneous agents, as forcing a single global mean and variance to scale observations from different local distributions disrupts normalization, leading to inconsistent updates and unstable convergence. This confirms that environmental heterogeneity undermines the effectiveness of shared normalization and highlights the need for locally personalized parameters.

\section{Conclusion}
This work tackled the critical challenge of handling heterogeneous input distributions in FedRL. 
We introduced a PON approach, which dynamically updates and applies mean and variance statistics for each agent. Our experiments on three MuJoCo tasks demonstrated that PON accelerates training and improves final performance compared to independent learning and standard FedRL baselines. Furthermore, our investigation revealed that sharing normalization parameters across agents fails in heterogeneous environments, underlining the necessity of local personalization. Future research may explore integrating other personalized mechanisms, such as adaptive loss reweighting or environment-specific policy components, to enhance further FedRL robustness in large-scale and highly variable real-world scenarios.

\section*{Acknowledgment}
This work was supported in part by the National Science Foundation under Grant 2047010, Grant 2117822, Grant 2047064 and Grant 2320972; and in part by the Department of Transportation under Grant 69A3552348304.

\bibliographystyle{IEEEtran}
\bibliography{ref}
\vspace{12pt}

\end{document}